\documentclass[runningheads]{llncs}

\usepackage[T1]{fontenc}
\usepackage{graphicx}
\usepackage{comment}
\usepackage{amsmath,amssymb}
\usepackage{color}
\usepackage{url}
\usepackage{hyperref}
\usepackage{booktabs}
\usepackage{multirow}
\usepackage{pifont}
\usepackage{wrapfig}

\newif\ifreview
\reviewfalse

\ifreview
	\usepackage{lineno}

	\linenumbers
\fi

\begin{document}

%%%%%%%%%%%%%%%%%%%%% Add submission id, track, and title. %%%%%%%%%%%%%%%%%%%%%

% TODO: Please insert your submission number here
\def\SubNumber{108}

% TODO: Please uncomment the track this paper will be submitted to, comment all other lines
%\def\GCPRTrack{Main Track}
%\def\GCPRTrack{Special Track: Pattern recognition in the life and natural sciences}
%\def\GCPRTrack{Special Track: Photogrammetry and remote sensing}
%\def\GCPRTrack{Young Researcher's Forum}
\def\GCPRTrack{Fast Review Track}

% TODO: Replace with your title
\title{Local Spherical Harmonics Improve Skeleton-Based Hand Action Recognition}
% You can use \thanks for acknowledgment as in: 
%\title{Title\thanks{XXX}}
%Do not add any acknowledgment to the draft 
% version that is used for the review process.  

\ifreview
	% ANONYMOUS SUBMISSION FOR REVIEW
	% DO NOT MODIFY these for the draft version that is used for the review process.
	\titlerunning{GCPR 2023 Submission \SubNumber{}. CONFIDENTIAL REVIEW COPY.}
	\authorrunning{GCPR 2023 Submission \SubNumber{}. CONFIDENTIAL REVIEW COPY.}
	\author{GCPR 2023 - \GCPRTrack{}}
	\institute{Paper ID \SubNumber}
\else
	% CAMERA READY SUBMISSION
	\titlerunning{Local Spherical Harmonics Improve Hand Action Recognition}
	% If the paper title is too long for the running head, you can set
	% an abbreviated paper title here

        \author{Katharina Prasse\inst{1}\orcidID{0009-0003-9502-1313} \and \\
	Steffen Jung\inst{1,2}\orcidID{0000-0001-8021-791X} \and
        Yuxuan Zhou\inst{3}\orcidID{0000-0002-7688-803X} \and
	Margret Keuper\inst{1,2}\orcidID{0000-0002-8437-7993}}
	
	\authorrunning{K. Prasse et al.}
	% First names are abbreviated in the running head.
	% If there are more than two authors, 'et al.' is used.

    	 \institute{University of Siegen, 57076 Siegen, Germany 
  \email{katharina.prasse@uni-siegen.de}\\
  \and Max Planck Institute for Informatics, Saarland Informatics Campus, Germany
	\and University of Mannheim, 68131 Mannheim, Germany}
 
\fi

\maketitle              % typeset the header of the contribution

\begin{abstract}
    Hand action recognition is essential. Communication, human-robot interactions, and gesture control are dependent on it.
    Skeleton-based action recognition traditionally includes hands, which belong to the classes which remain challenging to correctly recognize to date. 
    We propose a method specifically designed for hand action recognition which uses relative angular embeddings and local Spherical Harmonics to create novel hand representations.
    The use of Spherical Harmonics creates rotation-invariant representations which make hand action recognition even more robust against inter-subject differences and viewpoint changes. 
    We conduct extensive experiments on the hand joints in the First-Person Hand Action Benchmark with RGB-D Videos and 3D Hand Pose Annotations, and on the NTU RGB+D 120 dataset, demonstrating the benefit of using Local Spherical Harmonics Representations. 
    Our code is available at \url{https://github.com/KathPra/LSHR_LSHT}.

\end{abstract}
\section{Introduction}
Hand actions are everywhere. 
They can be observed during conversations, and provide valuable information about the atmosphere, hierarchy, backgrounds, and emotions.
Furthermore, their fine-grained differences make hand actions a challenging recognition task.
Skeleton-based action recognition naturally contains hand actions, which remain challenging to distinguish between.
Especially human-computer interactions require the accurate recognition of hand actions in order to understand e.g.~air quotes or the thumbs-up gesture.
When hand action recognition is further optimized, a vast field of applications stands to benefit from it; hand motion understanding in the medical field, gesture control, and robotics are mere examples.
Moreover, the increasing number of online interactions creates many scenarios when only a subset of the body joints is available for action recognition. 
Out of all body joints, hand joints are most likely to be included in digital recordings and have an expressive nature.
We thus focus on hand actions explicitly and our proposed method allows to better distinguish between them.
Since hands are very flexible, their embedding can benefit majorly from a local, relative description. 

\begin{figure}[t]
\begin{center}
\includegraphics[width=0.8\linewidth]{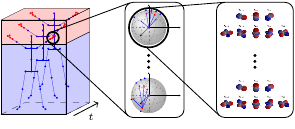}
\end{center}
\vspace{-0.5cm}
   \caption{Hands should receive particular emphasis as they contain the highest joint density and their mutual interaction is key to recognising hand actions. Hand joints (left) are first depicted as local spherical coordinates (middle) before being represented in terms of their Local Spherical Harmonics (right). The hand joint representation is then fed into the model as additional input.}
\label{fig:method}
\vspace{-0.3cm}
\end{figure}

Action recognition is an active field of research since the late 1990s \cite{minami1999real}. 
First advances into the field employed Recurrent Neural Networks [RNNs], among others Long Short-Term Memory Networks [LSTM], to capture the action over time \cite{du2015hierarchical,zhu2016co,zhang2017view}. 
Later, CNNs were fed action as a 2D map including semantic information as features e.g.,~joint type, time stamp, and position \cite{ke2017new,zhang2019view,weng2018deformable,liu2017enhanced}. 
In recent years, Graph Convolutional Networks [GCN] are frequently employed and show a strong performance \cite{si2018skeleton,yan2018spatial,adasgn2021iccv,chen2021channel}. In contrast to CNNs, GCNs can use the inherent structure of the skeleton data.
The joints are represented as nodes in the graph and the edges indicate their physical connections \cite{adasgn2021iccv} or correlations \cite{chen2021channel}. 
Action recognition uses various forms of input data, ranging from RGB+Depth video data to skeleton joint Cartesian coordinates extracted from videos. 
Using skeleton data increases model efficiency and enhances robustness against variations in viewpoint and appearance \cite{chen2021channel,adasgn2021iccv,zhang2020semantics}. 
In general, action recognition greatly benefits from understanding hand motion \cite{trivedi2021ntu,FirstPersonAction_CVPR2018}, which motivates us to focus particularly on hand joints. 
We describe each hand joint relative to the remaining hand joints. 
This way, the inter-joint relationships are explicitly included in the data and do not have to be inferred by the model. 

We propose to represent fine-grained motion through angular embeddings and local Spherical Harmonics, as visualized in \autoref{fig:method}.
We hypothesize that local spherical representations are better suited than Cartesian coordinates for several reasons: (1) Hand joints are very close in absolute position. 
When depicting them in terms of their relative position, small changes become more apparent. 
(2) The mutual interaction between fingers within each hand is one of the most important cues for action recognition. 
(3) Hands are very flexible. 
Considering their relative joint positions in terms of spherical projections facilitates robustness to slight motion variations.
In previous works, cylindrical \cite{weinland2006free} coordinates have been shown to improve detection accuracy.
We propose an angular embedding based on local Spherical Harmonics representations \cite{green2003spherical} as we find it suitable for describing the relative positions of the hand joints, i.e.~modelling inter-joint relationships.
Further, the data transformation makes the representation more robust against inter-subject differences and viewpoint changes, as it is partially rotation-invariant. 
We further propose hand joints' Local Spherical Harmonic Transforms \cite{green2003spherical}, a fully rotation-invariant representation.
To our knowledge, we are the first to employ Spherical Harmonics in the field of hand action recognition.

\noindent This study's main contributions can be summarized as follows:
\begin{enumerate}
\setlength\itemsep{0.1mm}
    \item We depict each hand joint in terms of its local neighbourhood to enable better hand-action recognition.
    \item We combine angular embeddings with the standard input as an explicit motion description to better represent the hand's inter-joint relationships.
    \item We combine Spherical Harmonic Transforms with the standard input as a rotation-invariant motion description to increase the robustness of hand action recognition against viewpoint or orientation changes.
    \item We show that angular representations, explicitly representing inter-joint relations, can improve the egocentric hand action recognition on FPHA \cite{FirstPersonAction_CVPR2018} by a significant margin.
    \item We further show that angular spherical embeddings can also be leveraged by classical human skeleton-based action recognition models such as CTR-GCN~\cite{chen2021channel}, improving hand-based action recognition accuracy.
    
\end{enumerate}

%-------------------------------------------------------------------------
\section{Related Work}
\subsection{Skeleton-Based Action Recognition}
Skeleton-based action recognition initially employed Recurrent Neural Networks (RNN) to learn features over time \cite{du2015hierarchical,zhu2016co,zhang2017view}. 
Long Short-Term Memory Networks (LSTMs) were popular as they were able to regulate learning over time. 
Furthermore, Temporal State-Space Models (TF) \cite{garcia2017transition}, Gram Matrices \cite{zhang2016efficient}, and Temporal Recurrent Models (HBRNN) \cite{du2015hierarchical} were popular choices. 
Prior to using learned features, hand-crafted features were employed and achieved competitive levels of accuracy compared to modern methods \cite{velivckovic2017graph,FirstPersonAction_CVPR2018}. 
Lastly, Key-pose models such as Moving Pose made use of a modified kNN clustering to detect actions \cite{zanfir2013moving}, whereas Hu et al. proposed a joint heterogeneous feature learning framework \cite{hu2015jointly}. 
Graph Convolution Networks (GCNs) for skeleton-based action recognition have been proposed e.g.,~in \cite{liu2020disentangling,2sagcn2019cvpr,tang2018deep,yan2018spatial,ye2020dynamic} and are defined directly on the graph, in contrast to Convolutional Neural Networks (CNNs). 
GCNs can be divided into two categories, i.e.~spectral \cite{bruna2013spectral,defferrard2016convolutional,kipf2016semi,henaff2015deep} and spatial \cite{yan2018spatial,chen2021channel} methods. 
While the research is mainly pushed in the direction of GCNs, CNNs and transformers remain part of the scientific discourse on action recognition \cite{duan2022revisiting,hu2022transrac,xu2022topology}.

While different methods have been proposed over time, one main tendency can be distilled from previous research: It is beneficial in terms of model accuracy to incorporate additional information \cite{zanfir2013moving,adasgn2021iccv,zhang2020semantics,chen2021channel,hu2015jointly,qin2022fusing}. Qin et al.~\cite{qin2022fusing} have included angles between several body joints into their model. Recently, multi-modal ensembles have been shown to increase the overall accuracy \cite{chen2021channel,chi2022infogcn,qin2022fusing}. 
While skeleton-based action recognition mainly focuses on the whole body, e.g.,~the popular NTU RGB-D 120 benchmark dataset [NTU 120] \cite{liu2019ntu}, several researchers have seen the merit in focusing on hand actions \cite{trivedi2021ntu,FirstPersonAction_CVPR2018}, and among them, Garcia et al.~published the First-Person Hand Action Benchmark [FPHA] \cite{FirstPersonAction_CVPR2018}. 
This dataset consists of egocentric recordings of daily hand actions in three categories, i.e.~\textit{social}, \textit{kitchen} and \textit{office actions}. Both datasets are included in this research. 
It is our goal to advance skeleton-based hand action recognition.
We thus limit the review of skeleton-based action literature to the most relevant works.

\subsection{Frequency Domain Representations for Action Recognition}
Representing the skeleton information in the frequency instead of the spatial domain has two main advantages. Firstly, noise can be removed more easily and secondly, rotation invariance can be introduced. When transferring data from the spatial to the frequency domain, the data is represented as a sum of frequencies. When the input's data format is Cartesian coordinates, a Fourier Transformation is commonly used; when the input is spherical coordinates, Spherical Harmonics can be employed. Besides the different input formats, these methods are equivalent. They take a function and map it from the spatial domain to the frequency domain by approximating it as sums of sinusoids \cite{brunton_kutz_2019}. Many previous works on action recognition have employed Fourier Transformation \cite{rodriguez2008action,weinland2006free,wang2012mining}.

Removing noise from human recording is always beneficial, e.g.~when a waving hand is shaking, excluding the shaking, a high-frequency motion, makes it easier to correctly identify the action.
Since the frequency domain data is complex, it has to be transformed in order to be fed into standard neural networks. 
Using the magnitude, a representation invariant to rotations, renders data normalization superfluous \cite{temerinac2007invariant} and can improve recognition accuracies in use cases with varying viewpoints. 

\subsection{Spherical Harmonics in 3D Point Clouds}
We are the first to employ Spherical Harmonics in skeleton-based action recognition to our knowledge. In 3D Point Clouds however, Spherical Harmonics are explored to make data representations robust or equivariant to rotations \cite{esteves2018learning,poulenard2019effective,spezialetti2020learning}. Esteves et al.~\cite{esteves2018learning} map 3D point clouds to spherical functions and propose spherical convolutions, which are robust against random rotations. Similarly, Spezialetti et al.~propose a self-supervised learning framework for using spherical CNNs to learn objects' canonical surface orientation in order to detect them independent of SO(3) rotations \cite{spezialetti2020learning}. Poulard et al.~\cite{poulenard2019effective} propose spherical kernels for convolution which directly operate in the point clouds. Li et al.~\cite{li2021closer} compare several aforementioned methods \cite{esteves2018learning,poulenard2019effective,spezialetti2020learning} in settings where the data points are either rotated around one or all three axes during training. They highlight the strong performance of SPH-Net \cite{poulenard2019effective} whose accuracy remains on the same level independent of the number of axis rotated around during training. Fang et al.~\cite{fang2020rotpredictor} points out the superiority of common CNNs over Spherical CNNs. Hence, we employ Spherical Harmonics as embeddings to which standard CNNs or GCNs can be applied.

%-------------------------------------------------------------------------
\section{Method}
We propose multi-modal hand joint representations from which our model learns feature representations.
Skeleton-based action recognition generally uses the five-dimensional Cartesian coordinates $X \in \mathbb{R}^{N \times M \times C \times T \times V}$, where $N$ describes the batch size, $M$ the number of persons in the action recording, $C$ the channel dimension, $T$ the number of frames, and $V$ the number of joints. 
We combine the Cartesian coordinates with a local embedding, created using Spherical Harmonics basis functions. % in increase hand action recognition accuracy.

\subsection{Local Spherical Coordinates}
\label{ssec:lsc}
\begin{figure}[ht]
\centering
\includegraphics[width=0.7\textwidth]{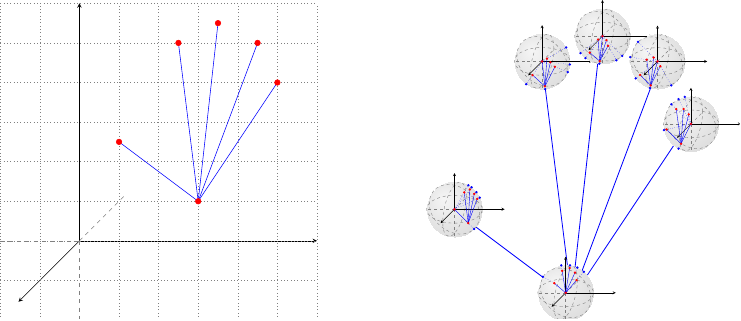}
\caption{Conversion between Cartesian global coordinates (left) and local spherical coordinates (right), where all coordinates are computed relative to each other joint. The global coordinates are Cartesian, while the local coordinates are spherical.}
\label{fig:glocal_hand}
\end{figure}

We first incorporate the hand joints' local neighbourhood by using each joint as the center of the coordinate system once while depicting the other joints' position relative to the center joint, as shown in \autoref{fig:glocal_hand}. 
We combine the local representations of all hand joints to capture fine-grained motion e.g.,~the differences between "making ok sign" and "making peace sign", even though the absolute positions of the thumb and index finger are very similar between both actions.

This transforms $X \in \mathbb{R}^{N \times M \times C \times T \times V}$ into $X_{loc} \in \mathbb{R}^{N \times M \times C \times T \times V \times V}$ where the last dimension contains the joint's local neighbourhood relative to its center joint's position. 
We compute the local neighbourhood for all hand joints, as we expect to gain the biggest advantage from investigating their local neighbourhood.

Moreover, we convert the local coordinates from Cartesian to Spherical coordinates as their angles are more suitable for describing hand motion than positions.
Spherical coordinates consist of three values ($r, \theta, \phi$) and are used to describe point positions in 3D space, as visualized in \autoref{fig:spher_vis} (a). They can be computed using Cartesian coordinates as inputs. 
The radius $r \in [0,\infty)$ is defined as the length of the vector from the origin to the coordinate point, i.e.~$r = \sqrt{x^2 + y^2 + z^2}$. 
The polar angle $\theta = \arctan \frac{ \sqrt{x^2 + y^2} }{z}$ describes the point's latitude and is defined for $\theta \in [0, \pi]$. 
The azimuthal angle $\phi = \arctan \frac{y}{x}$ describes the point's longitude and is defined for $\phi \in [0, 2\pi]$. 
The azimuth describes the point's location in the xy-plane relative to the positive x-axis.

\subsection{Spherical Harmonics based Hand Joint Representations}
\label{ssec:SHT}

We propose angular embeddings and Spherical Harmonics as novel representations for hand action recognition, which we realize through local spherical coordinates.
All Spherical Harmonics-based representations naturally offer a coarse to fine representation while being interpretable in terms of frequency bands developed on a sphere.
This includes representations using the Spherical Harmonics basis functions and the full Spherical Harmonic Transform.
Moreover, the magnitude of Spherical Harmonic Transforms is rotation-invariant which is a helpful property for hand action recognition \cite{green2003spherical}.

\begin{figure}[ht]
\vspace{-0.6cm}
\begin{tabular}{p{0.3cm} c p{0.4cm} c}
\parbox[t][][c]{0.3cm}{ \hspace*{0.3cm}}&
\parbox[t][][c]{3.3cm}{ \ \ \ \includegraphics[width=0.25\textwidth]{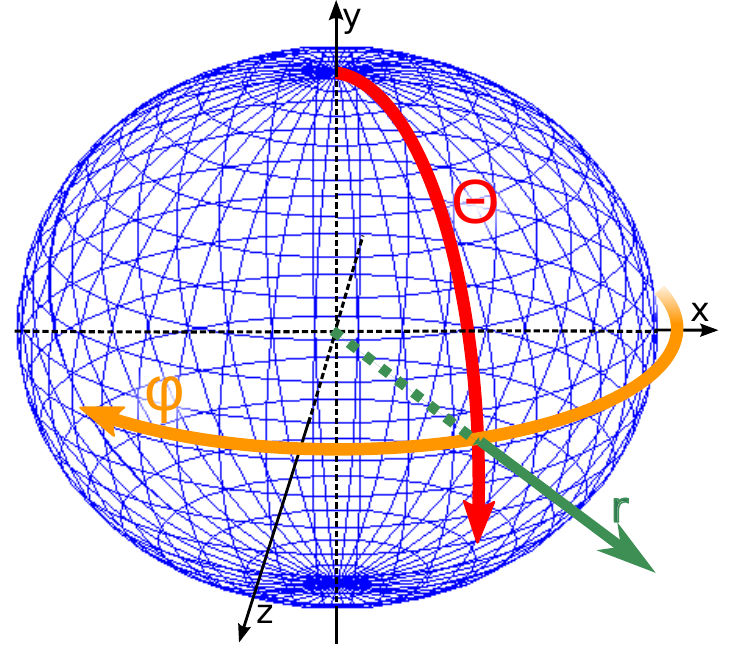}}&
\parbox[c][][c]{0.4cm}{ \hspace*{0.4cm}}&
\parbox[t][][c]{7cm}{\includegraphics[width=0.57\textwidth]{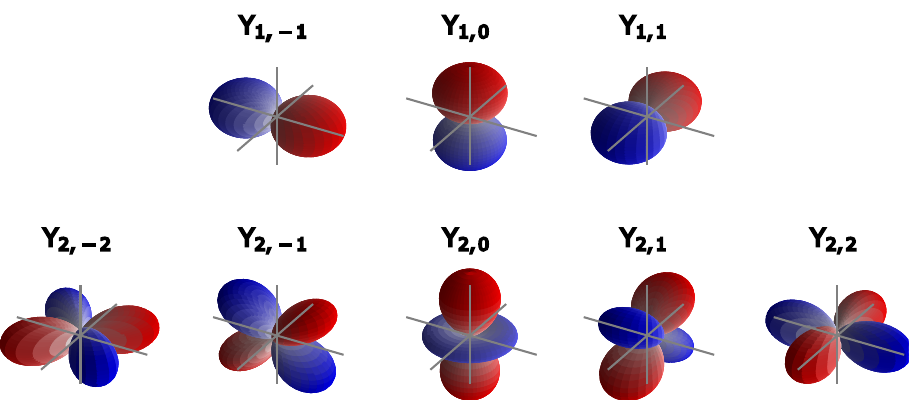}}\\
&&&\\
&(a)&& \  \  (b)\\
\end{tabular}
\caption{(a) Conversion between Cartesian coordinates ($x,y,z$) and spherical coordinates ($r, \theta, \phi$). Spherical coordinates consist of the radius \textit{r}, the polar angle \textit{$\theta$} and the azimuthal angle \textit{$\phi$}; (b) Visualization of the real part of Spherical Harmonics, where red indicates positive values while blue indicates negative values. The distance from the origin visualizes the magnitude of the Spherical Harmonics in the respective angular direction.}
\vspace{-0.2cm}
\label{fig:spher_vis}
\end{figure}

\autoref{fig:spher_vis} visualizes how we first transform the hand coordinates from Cartesian to spherical coordinates (a) before representing them in terms of their Spherical Harmonic basis functions (b). 

A function on the sphere can be represented in Spherical Harmonics as 

\begin{equation}
\label{eq:orth_exp}
 f(\theta, \phi) = \sum^{\infty}_{\ell = 0} \sum^{m=\ell}_{m=-\ell} a^{\ell}_{m} Y^{\ell}_{m}(\theta, \phi)
\end{equation}
with
\begin{equation}
\label{eq:SHT}
 Y_{\ell,m}(\theta, \phi) = \sqrt{\frac{(\ell-m)!(2\ell+1)}{(\ell+m)! 4\pi}} {e}^{im\phi} P^m_\ell cos\theta
\end{equation}
where $Y^{\ell}_{m}(\theta, \phi)$ are the Spherical Harmonics basis functions and the \textit{!} indicates a factorial operation.
Spherical Harmonics take two spherical coordinates as input, the azimuth $\theta$ and the polar angle $\phi$. 
They further include the associated Legendre polynomial $P^m_\ell$ and have the parameter degree $\ell$, in accordance to which the order $m$ is set, i.e.~$m \in [-\ell,\ell]$; both $\ell$ and $m$ are real numbers. 
The magnitude of the Local Spherical Harmonics basis function Representation [LSHR] is invariant against rotations around the y-axis, since the magnitude of the complex exponential function is $1$. 
The magnitude of the Local Spherical Harmonic Transforms [LSHT] is SO(3) rotation invariant.
We argue that such representations are beneficial when learning to recognize hand actions across different views.
We further expect these representations to support recognition in spite of hand orientation differences between subjects.

The Spherical Harmonics for $\ell \in \{1,2\}$ are employed for skeleton-based hand action recognition, as visualized in \autoref{fig:spher_vis} (b).
The inclusion of $\ell=0$ does not contain action-specific information, and we assume that all relevant information is contained within the first two bands, i.e. $\ell \leq 2$. 
The parameter $\ell$ is defined as the number of nodal lines, or bands, thus with larger $\ell$, higher frequencies are represented; the removal of high frequencies reduces the noise in the data. 

The chosen Spherical Harmonics representation is stacked along the channel dimension and concatenated to the original input.
Since Spherical Harmonics are complex numbers, i.e. $ x = a + b i$, they cannot directly be fed into standard neural networks.
They need to either be represented by their real and imaginary parts, or by their magnitude and phase. 
When using a single part, the representation is no longer complete and the input cannot be recovered entirely. 
The real and imaginary parts are a complete representation of the complex number. When only the real or only the imaginary part is used, the representation is no longer complete, mathematically speaking.
The magnitude has the property of being rotation-invariant \cite{green2003spherical}.

\subsection{Models}
We include angular embeddings, a local Spherical Harmonics representation (LSHR), and Spherical Harmonic Transforms (LSHT) both in a simple baseline model (GCN-BL) and an advanced model (CTR-GCN), both proposed by Chen et al.~\cite{chen2021channel}. 
The Channel-wise Topology Refinement Graph Convolution Model (CTR-GCN), proposed by Chen et al., achieves state-of-the-art results with a clean model architecture and a small number of epochs \cite{chen2021channel}. 
Both models are described in the subsequent paragraph, closely following Chen et al.'s description. 
They each consist of 10 layers, where each layer contains both a spatial graph convolutional network (GCN) module and a temporal convolutional network (TCN) module. 
While the CTR-GCN model learns a non-shared topology for the channels and dynamically infers joint relations during inference, the GCN-BL model learns a static topology shared between all channels. 
Further implementation details can be found in the supplementary material \autoref{seca:impli}.

\subsection{Evaluation}

The angular embeddings and Spherical Harmonics are evaluated in terms of their accuracy improvement both on all body joints [Imp.] and with a focus on hand joints [Hand Imp.]. During training, we randomly rotate the input data.

The local Spherical Harmonics representations are concatenated to the standard model input, the Cartesian coordinates, before the first layer. This causes a negligible increase in the number of parameters since only the input dimensionality is modified but not the layer outputs. When evaluating the effect of hand joints' angular embeddings or Spherical Harmonics, the model is compared to the original implementation. In order to maintain the data structure of $ X \in \mathbb{R}^{N \times M \times C \times T \times V}$, zeros are inserted for all non-hand joints. 
Our ablation studies contain a model trained exclusively on angular embeddings evaluated on FPHA \cite{FirstPersonAction_CVPR2018}. Furthermore, the NTU120 \cite{liu2019ntu} ablations include a baseline model of identical dimensionality, where random numbers replace the angular embeddings.

%-------------------------------------------------------------------------
\section{Experiments}
\subsection{Datasets}
The benefits of Local Spherical Harmonics representations (LSHR) and Local Spherical Harmonic Transforms (LSHT) of hand joints are shown on two datasets: Mainly, the First-Person Hand Action Benchmark is assessed. 
\begin{wrapfigure}{r}{5cm}
\vspace{-0.7cm}
\begin{center}
   \includegraphics[width=2cm, height=2.5cm]{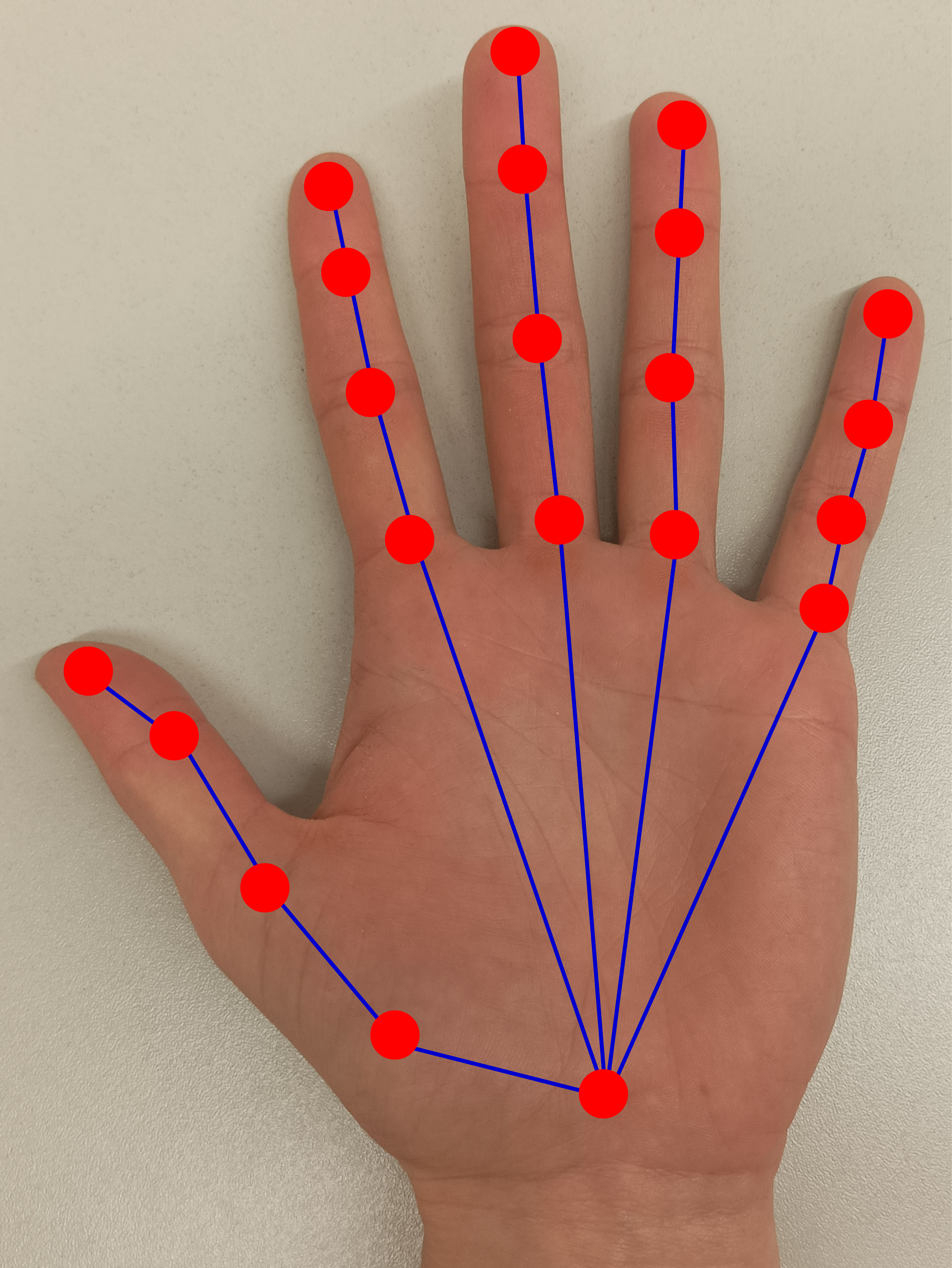}
   \hspace{0.2cm}
   \includegraphics[width=2cm, height=2.5cm]{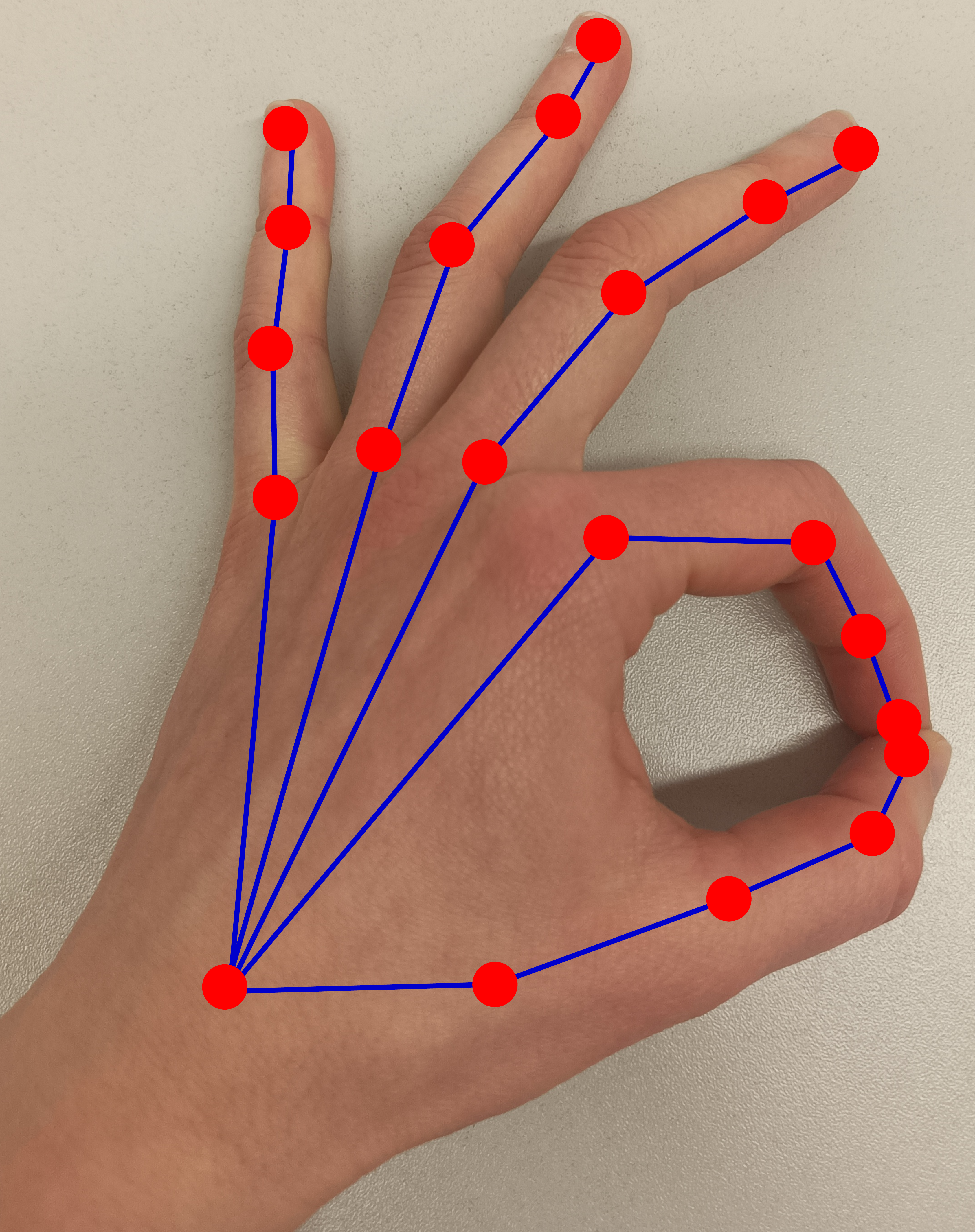}
\end{center}
\vspace{-1em}
   \caption{Visualization of hand joints in First-Person Hand Action Benchmark \cite{FirstPersonAction_CVPR2018} (own visualization).}
   \vspace{-0.5cm}
\label{fig:fpha}
\end{wrapfigure}
It is created by Garcia-Hernando et al. \cite{FirstPersonAction_CVPR2018} and  consists of shoulder-mounted camera recordings of six subjects each performing 45 action classes four to six times. The dataset contains 1175 skeleton recordings over time, including the wrist and four joints for each finger as shown in \autoref{fig:fpha}. This dataset is split 1:1 into train and test sets. 
The viewpoints differ between recordings, which makes the recognition simultaneously Cross-Subject and Cross-Setup tasks.

Secondly, the benchmark dataset NTU RGB+D 120 \cite{liu2019ntu} is employed. 
It contains 120 action classes, which can be split into daily, medical, and criminal actions \cite{liu2019ntu}. The dataset contains 114,480 videos featuring 106 distinct subjects in 32 scenarios (camera height and distance to subject). The dataset can be evaluated in two settings, i.e.~Cross-Subject and Cross-Setup. In the Cross-Subject evaluation, the 106 subjects are split into the train and test set. In the Cross-Setup setting, some recording angles and backgrounds are exclusively used for training, while others are used for testing. Both cases have the same train-test split, i.e.~1:1. The computation of local Spherical Harmonics was exclusively done for the eight hand joints, i.e.~both left and right wrist, hand, thumb, and finger.

\subsection{Implementation Details}
Experiments are conducted on NVIDIA Tesla V100 GPUs with the PyTorch learning framework. The models proposed by Chen et al.~\cite{chen2021channel}, GCN-BL and CTR-GCN, have the same hyperparameter settings, i.e. SGD with a momentum of 0.9, weight decay of 0.0004, and training for 65 epochs including 5 warm-up epochs. The learning rate is initialized with 0.1 and decays by a factor of 0.1 at epochs 35 and 55. The batch size is 64 and all action recordings are resized to 64 frames. 
For the FPHA dataset, a batch size of 25 was chosen, due to the smaller size of this dataset. 
Experiments are run on AMD Ryzen 9 5900x.

\subsection{Experimental Results}

Experiments on the First-Person Hand Action Benchmark (FPHA) demonstrate the strength of angular embeddings, a local Spherical Harmonics representation [LSHR], as shown in \autoref{table:results_FPHA_CTRGCN_BL_MOD}. 
The model accuracy increases for both the GCN-BL \cite{chen2021channel} and the CTR-GCN \cite{chen2021channel} model. We ablate our method by selecting angular embeddings as the sole model input; they outperform the original model by a large margin. The original CTR-GCN model suffers from severe overfitting causing their test accuracy to remain below the GCN-BL model's test accuracy. The original GCN-BL model overfits less and thus has a higher accuracy.

\begin{table}[h!]
%\vspace{-0.3cm}
\footnotesize
    \centering
    %\begin{tabular}{|l|c|c|c|}
    \begin{tabular}{l@{\hspace{1.4cm}}c@{\hspace{1.2cm}}c@{\hspace{1.2cm}}c}
        \toprule
        \multirow{2}{*}{\textbf{Method}} & \multirow{2}{*}{\textbf{Param.}} & \textbf{Rotation} & \multirow{2}{*}{\textbf{Acc (\%)}} \\
        &&\textbf{Invariance}&\\
        \midrule
        GCN BL & 2.1 M & \ding{55} & 80.52 \\
        + LSHR (R\&I) & 2.1 M & \ding{55} & 88.35 \\
        + \textbf{LSHR (M)} & 2.1 M & \ding{51} & \textbf{89.04} \\
        + LSHT (M) & 2.1 M & \ding{51} & 87.30 \\
        excl. LSHR (M) & 2.1 M & \ding{51} & 83.83 \\
        \midrule
        CTR-GCN & 1.4 M & \ding{55} & 74.26 \\
        + LSHR (R\&I) & 1.5 M & \ding{55} & 90.26 \\ 
        + \textbf{LSHR (M)} & 1.5 M & \ding{51} & \textbf{92.52} \\
        + LSHT (M) & 1.4 M & \ding{51} & 89.04 \\
        excl. LSHR (M) & 1.5 M  & \ding{51} & 85.57 \\
        \bottomrule
    \end{tabular}
    \vspace{0.7cm}
    \caption{Evaluation of the GCN-BL \cite{chen2021channel} and CTR-GCN \cite{chen2021channel} Model with Local Spherical Harmonics Representations [LSHR] and Local Spherical Harmonic Transforms [LSHT] evaluated on FPHA \cite{FirstPersonAction_CVPR2018} using the real and imaginary parts (R\& I) or the rotation-invariant magnitude (M), compared against the original model, and LSHR exclusively (*). Our Method significantly increases accuracy.}
    \vspace{-0.7cm}
    \label{table:results_FPHA_CTRGCN_BL_MOD}
\end{table}

\newpage
As \autoref{table:results_FPHA_CTRGCN_BL_MOD} shows, the largest increase over the original model is achieved with the CTR-GCN model, using the rotation-invariant magnitude of the angular embedding, a local Spherical Harmonics representation [LSHR] (+18\%). 
Local Spherical Harmonic Transforms [LSHT] also outperform the original model by a large margin (+15\%).
The results indicate that learning a local relative hand representation directly from the basis functions is favourable over employing their Spherical Harmonics.
The performance of the GCN-BL model differs from the CTR-GCN model, possibly due to the design of the model. 
While GCN-BL has a shared topology, the CTR-GCN model uses a channel-wise topology. 
Further angular embedding formats are reported in the supplementary material \autoref{seca:fpha}. 

When comparing our method to others evaluated on the first-person hand action benchmark, we clearly outperform previous models by a large margin, as shown in \autoref{table:comp_FPHA}. Our accuracy is 7\% higher than the best previously reported model using the Gram Matrix \cite{zhang2016efficient}.

\begin{wraptable}{p}{0.45\textwidth}
\vspace{-1.2cm}
\footnotesize
\begin{center}
\begin{tabular}{|l|c|}
\hline
Method & Acc (\%) \\
\hline\hline
1-layer LSTM \cite{FirstPersonAction_CVPR2018} & 78.73 \\
2-layer LSTM \cite{FirstPersonAction_CVPR2018} & 80.14 \\
\hline
Moving Pose \cite{zanfir2013moving} & 56.34 \\
Lie Group \cite{vemulapalli2014human} & 82.69\\
HBRNN \cite{du2015hierarchical} & 77.40\\
Gram Matrix \cite{zhang2016efficient} & \underline{85.39} \\
TF \cite{garcia2017transition} & 80.69\\
JOULE-pose \cite{hu2015jointly}& 74.60\\
TCN \cite{kim2017interpretable} & 78.57\\
LEML \cite{huang2015log} & 79.48 \\
SPDML-AIM \cite{harandi2017dimensionality} & 78.40 \\
SPDNet \cite{huang2017riemannian} & 83.79\\
SymNet-v1 \cite{wang2021symnet} & 81.04\\
SymNet-v2 \cite{wang2021symnet} & 82.96\\
\hline
GCN-BL \cite{chen2021channel} & 80.52 \\
CTR-GCN \cite{chen2021channel} & 74.26 \\
\textbf{Ours} & \textbf{92.52}\\
\hline
\end{tabular}
\end{center}
\vspace{-0.5em}
\caption{Model Evaluation on First-Person Hand Action Benchmark \cite{FirstPersonAction_CVPR2018}. Our model using the magnitude of the angular embeddings outperforms all other models.}
\vspace{-0.5cm}
\label{table:comp_FPHA}
\end{wraptable}

Our method is clearly superior when evaluated on a hands-only dataset such as FHPH \cite{FirstPersonAction_CVPR2018}. 
Furthermore, when evaluating it on a body dataset, it remains superior to the original model's performance.
We evaluate angular embeddings, a local Spherical Harmonics representation [LSHR], on the NTU120 dataset. For both the Cross-Subject and Cross-Setup benchmarks, we assess various input formats. 
These benchmarks give further insights into the performance of rotation-invariant features.
Within the Cross-Subject benchmark, we can evaluate the robustness of angular embeddings and Spherical Harmonics against inter-subject differences.
The Cross-Setup benchmark allows us to evaluate the rotation-invariance of our method explicitly.
The results using the GCN-BL model confirm the findings from the FPHA dataset, that angular embeddings, a local Spherical Harmonics representation [LSHR], increase the overall model accuracy (see \autoref{table:results_NTU120_CTRGCNBL}). 
This improvement is even larger when exclusively assessing hand-related action classes [Hand Imp.].
The full list of hand vs. non-hand-related action classes can be found in the supplementary material \autoref{seca:handclass}. 
In the Cross-Subject setting, the full spectrum, i.e.~real and imaginary parts of angular embeddings, induces the largest accuracy increase, as expected. In this case, the use of a rotation-invariant representation is not as advantageous for cross-subject action recognition. 
In the Cross-Setup setting, the largest accuracy increase was achieved when using the magnitude of the angular embedding, in line with our expectations, that the rotation invariant representation is well suited for this setting. 
The phase information reduces model accuracy, as it contains rotation information and thus hinders recognition when different viewpoints are compared. 
\begin{table}[ht]
\vspace*{-0.3cm}
\footnotesize
\begin{center}
\begin{tabular}{|l|l|c|c|c|c|}
\hline
\multirow{2}{*}{Dataset} & \multirow{2}{*}{Format} & Rand. BL & Ours & Imp. & Hand Imp. \\
& & Acc. (\%)&Acc. (\%)& &\\
\hline\hline
\multirow{5}{*}{X-Sub} & Real & 83.89 & 84.38 & $\uparrow +0.5$&$\uparrow +0.9$\\ 
 & Imaginary & 83.89 & 84.39 & $\uparrow +0.5$&\pmb{$\uparrow +1.0$}\\ 
 & Magnitude & 83.89 & 84.20 & $\uparrow +0.3$&$\uparrow +0.3$\\ 
 & Real \& Imag. & 83.18 & 84.01 & \pmb{$\uparrow +0.8$}&\pmb{$\uparrow +1.0$}\\ 
 & Mag. \& Phase & 83.18 & 83.72 & $\uparrow +0.5$&$\uparrow +0.7$\\
\hline
\multirow{5}{*}{X-Set} & Real & 85.62 & 85.93 & $\uparrow +0.3$&$\uparrow +0.5$\\ 
 & Imaginary & 85.62 & 85.98 & $\uparrow +0.4$&$\uparrow +0.7$\\ 
 & Magnitude & 85.62 & 86.21 & \pmb{$\uparrow +0.6$}& \pmb{$\uparrow +1.0$} \\ 
 & Real \& Imag. & 85.49 & 85.91 & $\uparrow +0.4$&$\uparrow +0.7$\\  
 & Mag. \& Phase & 85.49 & 85.39 & $\downarrow -0.1$&$\downarrow -0.5$\\
\hline
\end{tabular}
\end{center}
\caption{Format Comparison of the joint modality using GCN-BL with angular embeddings on NTU120. Our method increases overall accuracy (Imp.) and hand-related action accuracy (Hand Imp.) compared to a random baseline (Rand. BL).}
\label{table:results_NTU120_CTRGCNBL}
\vspace*{-0.9cm}
\end{table} 
When investigating the differences between local angular embeddings, a local Spherical Harmonics representation [LSHR], and Spherical Harmonic Transforms [LSHT] on the NTU120 dataset, it becomes apparent, that the GCN-BL model benefits from the inclusion of Local angular Spherical Harmonics Representations [LSHR].
The CTR-GCN model, however, achieves higher levels of accuracy when the local Spherical Harmonic Transforms [LSHT] are employed. 
\begin{table}[ht]
\vspace{-0.3cm}
\footnotesize
\begin{center}
\begin{tabular}{|l|l|l|c|c|c|}
\hline
\multirow{3}{*}{Dataset} & \multirow{3}{*}{Model} & \multirow{3}{*}{Joint} & \multirow{1.8}{*}{Original} & LSHR & LSHT \\ 
&&& \multirow{2}{*}{Acc. (\%)} & (Ours) & (Ours)\\ 
&& & &Acc. (\%)&Acc. (\%) \\ 
\hline\hline
\multirow{4}{*}{X-Sub} & \multirow{2}{*}{GCN-BL} & Loc. & 83.75 & 84.20 \pmb{($\uparrow 0.5$)} & 84.03 ($\uparrow 0.3$ \\ 
& & Vel. & 80.30 & 80.53 \pmb{($\uparrow 0.2$)} & 80.32 ($\pm 0$) \\
\cline{2-6}
& \multirow{2}{*}{CTR-GCN} & Loc. & 85.08 & 85.31 \pmb{($\uparrow 0.2$)} & 85.27 \pmb{($\uparrow 0.2$)} \\ 
& & Vel. & 81.12 & 81.45 ($\uparrow 0.3$) & 81.53 \pmb{($\uparrow 0.4$)} \\ 
\hline
\multirow{4}{*}{X-Set} &\multirow{2}{*}{GCN-BL} & Loc. & 85.64 & 86.21 \pmb{($\uparrow 0.6$)} & 85.75 ($\uparrow 0.1$)\\ 
& & Vel. & 82.25 & 82.41 \pmb{($\uparrow 0.2$)} & 82.25 ($\pm 0$)\\ 
\cline{2-6}
&\multirow{2}{*}{CTR-GCN} & Loc. & 86.76 & 86.63 ($\downarrow 0.1$) & 87.01 \pmb{($\uparrow 0.3$)}\\ 
& & Vel. & 83.12 & 83.32 ($\uparrow 0.2$) & 83.42 \pmb{($\uparrow 0.3$)} \\
\hline
\end{tabular}
\end{center}
\caption{Single modality evaluation using local Spherical Harmonics representations [LSHR], and rotation-invariant Local Spherical Harmonic Transforms [LSHT] on both NTU120 benchmarks \cite{liu2019ntu}. Rotation-invariant hand joint representations increase the model's accuracy.}
\label{table:results_fullharmonics}
\vspace{-0.9cm}
\end{table} 

The ensemble performance is increased by the inclusion of angular embeddings, local Spherical Harmonics representations [LSHR], or Spherical Harmonic Transforms [LSHT] as reported in \autoref{table:ctrgcn_ensemble}. 
The model using the full Local Spherical Harmonic Transforms [LSHT] outperforms the other models by a small margin.
As our method only includes joints, the bone modalities are always taken from the original model and do not contain any angular embeddings or Spherical Harmonic Transforms.

\begin{table}[ht]
\begin{center}
\begin{tabular}{l@{\hspace{0.9cm}}l@{\hspace{0.8cm}}c@{\hspace{0.8cm}}c}
 %\hline
 \toprule
\multirow{ 2}{*}{\textbf{Modality}} & \multirow{ 2}{*}{\textbf{Method}} & \multicolumn{2}{c}{\textbf{NTU RGB+D 120}} \\ 
  && \textbf{X-Sub (\%)} & \textbf{X-Set(\%)}\\
%\hline\hline
\midrule
 \multirow{3}{*}{Loc.} & CTR-GCN &\textbf{88.8} & 90.0 \\
 & LSHR (full spectrum) & 88.6 & 90.1\\
 & LSHR (rotation invariant) & \textbf{88.8} & 90.0 \\
 & LSHT (rotation invariant) & \textbf{88.8} & \textbf{90.2} \\
% \hline
\midrule
 \multirow{3}{*}{Loc. \& Vel.} & CTR-GCN & 89.1 & 90.6 \\
 & LSHR (full spectrum)& 88.9 & 90.5\\
 & LSHR (rotation invariant) & 89.1 & 90.6\\
& LSHT (rotation invariant) & \textbf{89.2} & \textbf{90.7}\\
%\hline
\bottomrule
\end{tabular}
\end{center}
\vspace{1cm}
\caption{Ensemble Evaluation of the CTR-GCN model in its original version compared with two local Spherical Harmonics representations [LSHR], full-spectrum (real and imaginary part) and rotation invariance (magnitude), and the full Spherical Harmonic Transforms [LSHT]. LSHT outperforms all other models by a small margin.} 
\label{table:ctrgcn_ensemble}
%\vspace{-1cm}
\end{table}

We conduct further experiments on the NTU120 benchmarks comparing rotation-invariant and complete angular embeddings, a local Spherical Harmonics representation [LSHR], for which we report both the hand accuracy and the overall accuracy in the supplementary material \autoref{seca:ntu120}. 
As our method focuses on hand joints, the hand accuracy increases are larger than the overall accuracy increases, when compared to the original model. 
%------------------------------------------------------------------------
\section{Discussion \& Conclusion}

Our rotation-invariant joint representations on the basis of local Spherical Harmonics improve skeleton-based hand action recognition. 
Their inclusion allows us to better distinguish between fine-grained hand actions. 
Both the theoretical framework and the experimental results affirm that hand joints' local Spherical Harmonics improve skeleton-based hand action recognition. 
We clearly outperform other methods the hands-only First-Person-Hand-Action-Benchmark \cite{FirstPersonAction_CVPR2018}.
The evaluations on the NTU120 Cross-Subject and Cross-Setup benchmarks further confirm our findings. 
Our experiments have shown that rotation-invariant hand representations increase robustness against inter-subject orientation differences and viewpoint changes, thus resulting in higher accuracy levels.
On the NTU120 dataset, the overall best model performance is achieved in the Cross-Setup setting using the joint location modality, as expected due to the rotation invariance of the features. 
The Cross-Subject setting benefited more from the rotation invariance than expected, hinting that inter-subject differences can be reduced using angular embeddings. 
Thinking of the action "making a peace sign" exemplifies this, as different subjects have slightly different orientations of their hands. 
We aim to open the door for future research in the adaptation of angular embedding to this and other data modalities. 
Furthermore, the number of hand joints included in the NTU120 is four per hand, which appears to make the sampling of additional data points a meaningful investigation.

\section*{Acknowledgements} 
This work is partially supported by the BMBF project 16DKWN027b Climate Visions and DFG research unit 5336 Learning2Sense. All experiments were run on the computational resources of the University of Siegen and the Max Planck Institute for Informatics. We would like to thank Julia Grabinski for her guidance.

\bibliographystyle{splncs04}
\bibliography{108-main.bib}

\newpage
\appendix
\chapter*{Supplementary Material}
\section{First Person Hand Action Benchmark}
\label{seca:fpha}

As part of our ablation, we evaluate our method on the First Person Hand Action Benchmark [FPHA]\cite{FirstPersonAction_CVPR2018}. Since angular embeddings are complex, we need to transform them prior to being able to use them in a standard neural network. Complex numbers can be represented either in terms of their real and imaginary part or in terms of their angle and magnitude. We report all input formats for the GCN-BL model in \autoref{table:results_FPHA_CTRGCN_BL} and for the CTR-GCN model in \autoref{table:results_FPHA_CTRGCN}. All formats outperform their respective baseline by a large margin.

\begin{table}[ht]
\vspace{-1em}
    \centering
    \begin{tabular}{|l|c|c|}
            \hline
            Method & Param. & Acc (\%) \\% & Best epoch \\
            \hline\hline
            \textit{GCN BL (baseline)} & \textit{2.1 M} & \textit{80.52} \\% & 65\\ 
            \hline
            + LSHR (R/I) & 2.2 M & 88.35 \\% & 54 \\ 
            + LSHR (M/P) & 2.2 M &  89.74\\% & 63 \\ 
            \hline
            + \textbf{LSHR (R)} & 2.1 M &  \textbf{89.91} \\% & 61\\
            + LSHR (I) & 2.1 M  & 89.04 \\% & 54\\
            + LSHR (M) & 2.1 M & 89.04 \\% & 54\\
            %+ LSHT (P) & 2.1 M & \ding{55}  & 86.95 \\% & 58\\
            \hline
            + spher. coord. angles & 2.1 M  & 87.83 \\% & 63\\ 
            \hline
            excl. LSHR (M) & 2.1 M  & 83.83* \\% & 63\\ 
            \hline
    \end{tabular}
    \vspace{0.3cm}
    \caption{Evaluation of GCN-BL Model with Local Spherical Harmonics Representation on First Person Hand Action Benchmark \cite{FirstPersonAction_CVPR2018} using various combinations of the real part (R), the imaginary part (I), the magnitude (M), and the phase (P), evaluated against the original model and LSHR exclusively (*).}
    \label{table:results_FPHA_CTRGCN_BL}
\vspace{-2em}
\end{table}

For the GCN-BL model, the inclusion of the real part has the strongest positive effect, i.e.~9\%, as reported in \autoref{table:results_FPHA_CTRGCN_BL}.
All other input formats improve by at least 7.8\%, while the performance differences between formats are relatively small. 
The model using angular embeddings exclusively outperforms the original model using Cartesian coordinates by more than 3\%.
While this accuracy increase is noteworthy, the model does benefit from containing Cartesian coordinates in combination with angular embeddings, as these formats' accuracy levels are up to 5\% higher. 
Similarly, the concatenation of the local spherical coordinates $(\theta, \phi)$ improves the model accuracy beyond the original level but remains below the angular embedding's performance (LSHR).

\begin{table}[ht]
\vspace{-1em}
    \centering
    \begin{tabular}{|l|c|c|}
            \hline
            Method & Param. & Acc (\%) \\% & Best epoch \\
            \hline\hline
            \textit{CTR-GCN (baseline) }& \textit{1.4 M} & \textit{74.26 }\\% & 65\\ 
            \hline
            + \textbf{LSHR (R/I)} & 1.6 M & \textbf{92.52} \\% & 54 \\ 
            + LSHR (M/P) & 1.6 M &  92.00 \\% & 63 \\ 
            \hline
            + LSHR (R) & 1.5 M & 89.65 \\% & 61\\
            + LSHR (I) & 1.5 M  & 90.26 \\% & 54\\
            + LSHR (M) & 1.5 M & 91.30 \\% & 54\\
            %+ LSHT (P) & 2.1 M & \ding{55}  & 86.95 \\% & 58\\
            \hline
            + spher. coord. angles & 1.4 M  & 89.91 \\% & 62\\ 
            \hline
            excl. LSHR (M) & 1.5 M  & 85.57* \\% & 63\\ 
            \hline
    \end{tabular}
    \vspace{0.3cm}
    \caption{Evaluation of CTR-GCN Model with Local Spherical Harmonics Representation on First Person Hand Action Benchmark \cite{FirstPersonAction_CVPR2018} using various combinations of the real part (R), the imaginary part (I), the magnitude (M), and the phase (P), evaluated against the original model and LSHR exclusively (*).}
    \label{table:results_FPHA_CTRGCN}
    \vspace{-2em}
\end{table}

Similar trends can be observed for the CTR-GCN model, as reported in \autoref{table:results_FPHA_CTRGCN}. 
It is interesting to note that for the CTR-GCN model, the accuracy level of the local spherical coordinates ($\theta, \phi$) is closer to the performance of angular embeddings, compared to the GCN-BL model. This indicates, that the more complex CTR-GCN model can partially learn the information contained in the angular embeddings, as their performance differs by a maximum of 2.5\%.

\section{NTU120}
\label{seca:ntu120}
To further strengthen the credibility of our method, we report the mean and confidence intervals over 5 random seeds (1,2,3,4,5) for the GCN-BL method using the joint modality on the Cross-Subject dataset. \autoref{table:gcnbl_meanstd} highlights the advantage of using angular embeddings for the rotation invariant angular embeddings and further confirms its superiority over the full spectrum as reported in the main paper. The magnitude of angular embeddings increases on average by $+0.4\%$ over the original model.

\begin{table}[ht]
\vspace{-1em}
\begin{center}
\begin{tabular}{|l|l|c|c|c|}
\hline
\multirow{3}{*}{Dataset} &\multirow{3}{*}{Model} & \multirow{1.8}{*}{Original} & Full spectrum & Rotation Invar. \\% & Best epoch \\
&&\multirow{2}{*}{Acc. (\%)} & (Ours) & (Ours) \\
&&  &Acc. (\%)&Acc. (\%)\\
\hline\hline
X-Sub & GCN-BL & $84.01 \pm0.2$ & $84.03 \pm0.2$ & $84.37 \pm0.1$\\
% CTR-GCN & $84.98 \pm 0.3$  & $85.03 \pm0.3$& $85.14 \pm0.1$ \\
\hline
\end{tabular}
\end{center}
\caption{Mean and standard deviation show the rotation invariant angular embeddings significantly outperform the GCN-BL model evaluated on NTU XSUB using the joint modality.}
\label{table:gcnbl_meanstd} 
\vspace{-2em}
\end{table} 

We have demonstrated the advantage of angular embeddings (LSHR) over local spherical coordinates ($\theta, \phi$) on the FPHA Dataset \cite{FirstPersonAction_CVPR2018} in \autoref{seca:fpha}. 
\autoref{table:input_ablation} shows that angular embeddings continuously outperform local spherical coordinates (LSC) also for the NTU120 Dataset \cite{liu2019ntu}. 
When comparing the original model and the local spherical coordinates, the performance slightly increases for the GCN-BL model and slightly decreases for the CTR-GCN model. Our method outperforms the baseline for both the Cross-Subject and the Cross-View setting, and both models.

\begin{table}[ht]
\vspace{-1em}
\begin{center}
\begin{tabular}{|l|l|c|c|c|c|}
\hline
\multirow{3}{*}{Dataset} & \multirow{3}{*}{Model} & \multirow{1.8}{*}{Original} & \multirow{1.8}{*}{LSC} & Full spectrum & Rotation Invar. \\% & Best epoch \\
&&\multirow{2}{*}{Acc. (\%)}& \multirow{2}{*}{Acc. (\%)} & (Ours) & (Ours) \\
&& &  &Acc. (\%)&Acc. (\%)\\
\hline\hline
\multirow{2}{*}{X-Sub} & GCN-BL & 83.75 & 83.95 ($\uparrow 0.2$) & 84.01 ($\uparrow 0.3$) & 84.20 \pmb{($\uparrow 0.5$)}\\
%& & 80.30 & & 80.59 \pmb{($\uparrow 0.3$)} & 80.53 ($\uparrow 0.2$)\\
%GCN BL & Bone Loc. & 84.82 & 85.22  \pmb{($\uparrow 0.5$)} & 84.09 ($\downarrow 0.7$)\\
%GCN BL & Bone Vel. & 80.15 & 80.48 \pmb{($\uparrow0.3$)}& 80.02 ($\downarrow0.1$) \\
%\cline{2-6}
& CTR-GCN & 85.08 & 85.01 ($\downarrow 0.1$) & 85.08 ( $\pm  0$)& 85.31 \pmb{($\uparrow 0.2$)}\\
%&  & 81.12& & 81.19 ($\uparrow  0.1$)& 81.45 \pmb{($\uparrow 0.3$)}\\
%CTR-GCN & Bone Loc. & 85.86 &  85.63 ($\downarrow 0.2$)& 85.72 ($\downarrow 0.1$)\\
%CTR-GCN & Bone Vel. & 81.46 & 81.08 ($\downarrow 0.3$)& 81.56  \pmb{($\uparrow 0.1$)}\\
\hline
\multirow{2}{*}{X-Set} &GCN-BL & 85.64 & 85.55 ($\uparrow 0.1$)& 85.91 ($\uparrow 0.3$) & 86.21 \pmb{($\uparrow 0.6$)} \\
%& & 82.25& & 82.54 \pmb{($\uparrow 0.3$)} & 82.41 ($\uparrow 0.2$)\\
%GCN BL & Bone Loc. & 86.38 & 86.08 ($\downarrow 0.3$) & 86.36 ($\pm 0$)\\
%GCN BL & Bone Vel. &  81.66 & 81.97 \pmb{($\uparrow 0.3$)} & 81.68 ($\pm 0$)\\
%\cline{2-6}
&CTR-GCN & 86.76& 86.65 ($\downarrow 0.1$) & 86.86 \pmb{($\uparrow 0.1$)} &  86.63 ($\downarrow 0.1$) \\
%& & 83.12 && 83.27 \pmb{($\uparrow 0.2$)} & 83.32 \pmb{($\uparrow 0.2$)}\\
%CTR-GCN & Bone Loc. & 87.52 & 87.37 ($\downarrow 0.2$) & 87.22 ($\downarrow 0.3$) \\
%CTR-GCN & Bone Vel. & 83.26 & 82.78 ($\downarrow 0.5$) & 83.07 ($\downarrow 0.2$)\\
\hline
\end{tabular}
\end{center}
\caption{Evaluation of local spherical coordinates (LSC) vs. angular embeddings (LSHR) using joint modality on NTU120 Dataset.}
\label{table:input_ablation}
\vspace{-2em}
\end{table} 

The evaluation of angular embeddings on NTU120 \cite{liu2019ntu} is given for all classes in the main paper.
In \autoref{table:results_NTU120_hands}, we report the hand accuracies, in line with the class categorization given in \autoref{seca:handclass}.
Our models' advantage is demonstrated the strongest when explicitly looking at hand action classes' performance. 
The accuracy on hand action classes increases when using the joint location and even more when using the joint velocity in the Cross-Subject data settings. 
In the Cross-View setting, the CTRGCN model's performance increases while the GCN-BL model's performance decreases when using joint velocity. It appears that the viewpoint change has a negative impact on the topology-shared GCN-BL, while CTR-GCN model's channel-wise topology can learn meaningful representations and increase its accuracy on hand action classes \cite{chen2021channel}.
 
\begin{table}[ht]
\vspace{-1em}
\begin{center}
\begin{tabular}{|l|l|l|c|c|c|}
\hline
\multirow{3}{*}{Dataset} & \multirow{3}{*}{Model} & \multirow{3}{*}{Modality} & \multirow{1.8}{*}{Original} & Full spectrum & Rotation Invar. \\% & Best epoch \\
&&& \multirow{2}{*}{Acc. (\%)} & (Ours) & (Ours) \\
&& &  &Acc. (\%)&Acc. (\%)\\
\hline\hline
\multirow{4}{*}{X-Sub} & \multirow{2}{*}{GCN-BL} & Joint Loc. & 74.33 & 74.54 ($\uparrow 0.2$) & 74.78 \pmb{($\uparrow 0.5$)}\\
& & Joint Vel. & 70.12 & 70.86 \pmb{($\uparrow 0.7$)} & 70.25 ($\uparrow 0.1$)\\
\cline{2-6}
& \multirow{2}{*}{CTR-GCN}  & Joint Loc. & 76.12 & 76.64 \pmb{($\uparrow 0.5$)}& 76.46 ($\uparrow 0.3$)\\
&  & Joint Vel. & 71.23 & 71.35 ($\uparrow  0.1$)& 71.88 \pmb{($\uparrow 0.7$)}\\
\hline
\multirow{4}{*}{X-Set} &\multirow{2}{*}{GCN-BL} & Joint Loc. & 77.60 & 78.21 ($\uparrow 0.6$) & 78.68 \pmb{($\uparrow 1.1$)} \\
& & Joint Vel. & 73.20 & 72.54 ($\downarrow 0.6$) & 73.14 \pmb{($\downarrow 0.1$)}\\
\cline{2-6}
&\multirow{2}{*}{CTR-GCN} & Joint Loc. & 79.17 & 79.65 \pmb{($\uparrow 0.5$)} &  79.23 ($\uparrow 0.1$) \\
& & Joint Vel. & 74.10 & 74.85 \pmb{($\uparrow 0.8$)} & 74.74 ($\uparrow 0.6$)\\
\hline
\end{tabular}
\end{center}
\caption{Single modality evaluation of GCN-BL model with Local Spherical Harmonics on NTU120 Dataset on hand-action classes. Our method shows an overall performance improvement.}%a strong performance in hand action classes.}
\label{table:results_NTU120_hands}
\vspace{-2em}
\end{table}

\autoref{table:results_NTU120} contains the same experiments as \autoref{table:results_NTU120_hands} and reports the overall accuracy levels. Experiments with the full-spectrum and the rotation invariant features assess their effect on various data modalities for both the Cross-Subject [X-Sub] and the Cross-Setup [X-Set] in comparison to the original model's accuracy.
Within the Cross-Subject benchmark, the overall highest accuracy increase is achieved when using the rotation-invariant magnitude of the embeddings for the joint modality, as shown in \autoref{table:results_NTU120}. For all data modalities, our model performs better or on par with the original models, i.e.,~GCN-BL \cite{chen2021channel} and CTR-GCN \cite{chen2021channel}. 
%Since the task on this dataset is to recognize actions performed by different subjects, the full spectrum was expected to outperform the rotation-invariant embedding. 
While the full spectrum increases the model accuracy for the simpler GCN-BL model, it does not have an effect on the advanced CTR-GCN model's performance. It appears that the rotation invariant features include meaningful information to better distinguish between actions, evening out inter-subject differences in the Cross-Subject benchmark.

In the Cross-Setup setting, the same performance trends are apparent for the GCN-BL model, as shown in \autoref{table:results_NTU120}. As expected, the joint location has the largest overall accuracy gain from the inclusion of rotation-invariant angular embedding, even more, when exclusively considering the hand class accuracy. For the joint velocity, the model accuracy increased for all cases, indicating that angular embeddings contain valuable information.
\begin{table}[ht]
\vspace{-1em}
\footnotesize
\begin{center}
\begin{tabular}{|l|l|l|c|c|c|}
\hline
\multirow{3}{*}{Dataset} & \multirow{3}{*}{Model} & \multirow{3}{*}{Modality} & \multirow{1.8}{*}{Original} & Full spectrum & Rotation Invar. \\% & Best epoch \\
&&& \multirow{2}{*}{Acc. (\%)} & (Ours) & (Ours) \\
&& &  &Acc. (\%)&Acc. (\%)\\
\hline\hline
\multirow{4}{*}{X-Sub} & \multirow{2}{*}{GCN-BL} & Joint Loc. & 83.75 & 84.01 ($\uparrow 0.3$) & 84.20 \pmb{($\uparrow 0.5$)}\\
& & Joint Vel. & 80.30 & 80.59 \pmb{($\uparrow 0.3$)} & 80.53 ($\uparrow 0.2$)\\
%GCN BL & Bone Loc. & 84.82 & 85.22  \pmb{($\uparrow 0.5$)} & 84.09 ($\downarrow 0.7$)\\
%GCN BL & Bone Vel. & 80.15 & 80.48 \pmb{($\uparrow0.3$)}& 80.02 ($\downarrow0.1$) \\
\cline{2-6}
& \multirow{2}{*}{CTR-GCN}  & Joint Loc. & 85.08 & 85.08 ( $\pm  0$)& 85.31 \pmb{($\uparrow 0.2$)}\\
&  & Joint Vel. & 81.12 & 81.19 ($\uparrow  0.1$)& 81.45 \pmb{($\uparrow 0.3$)}\\
%CTR-GCN & Bone Loc. & 85.86 &  85.63 ($\downarrow 0.2$)& 85.72 ($\downarrow 0.1$)\\
%CTR-GCN & Bone Vel. & 81.46 & 81.08 ($\downarrow 0.3$)& 81.56  \pmb{($\uparrow 0.1$)}\\
\hline
\multirow{4}{*}{X-Set} &\multirow{2}{*}{GCN-BL} & Joint Loc. & 85.64 & 85.91 ($\uparrow 0.3$) & 86.21 \pmb{($\uparrow 0.6$)} \\
& & Joint Vel. & 82.25 & 82.54 \pmb{($\uparrow 0.3$)} & 82.41 ($\uparrow 0.2$)\\
%GCN BL & Bone Loc. & 86.38 & 86.08 ($\downarrow 0.3$) & 86.36 ($\pm 0$)\\
%GCN BL & Bone Vel. &  81.66 & 81.97 \pmb{($\uparrow 0.3$)} & 81.68 ($\pm 0$)\\
\cline{2-6}
&\multirow{2}{*}{CTR-GCN} & Joint Loc. & 86.76 & 86.86 \pmb{($\uparrow 0.1$)} &  86.63 ($\downarrow 0.1$) \\
& & Joint Vel. & 83.12 & 83.27 \pmb{($\uparrow 0.2$)} & 83.32 \pmb{($\uparrow 0.2$)}\\
%CTR-GCN & Bone Loc. & 87.52 & 87.37 ($\downarrow 0.2$) & 87.22 ($\downarrow 0.3$) \\
%CTR-GCN & Bone Vel. & 83.26 & 82.78 ($\downarrow 0.5$) & 83.07 ($\downarrow 0.2$)\\
\hline
\end{tabular}
\end{center}
\caption{Single modality evaluation of Local Spherical Harmonics Representations [LSHR] on NTU120 \cite{liu2019ntu}. Our method shows a performance increase with angular embeddings.}
\label{table:results_NTU120}
\vspace{-2em}
\end{table} 

\autoref{table:ctrgcn_ri_quali} and \autoref{table:ctrgcn_mag_quali} exemplify our method in reporting individual class accuracy changes for the CTR-GCN model evaluated using the joint modality. 
Both tables contain all action classes whose accuracy changes by 4\% or more. 
For both models, the rotation invariant and the full spectrum one, a larger number of action classes' accuracy increases compared to the number of action classes with lower performance due to our method
The results clearly show that all improved action classes contain hand motion.
However, the class "play magic cube" (84) decreased in terms of model accuracy for both models. 
This indicates that the intra-subject differences are large in this action class and this action recognition is hindered rather than improved when including angular embeddings.
\begin{table}[ht]
\vspace{-1em}
\begin{center}
\begin{tabular}{|l|c|c|c|}
\hline
\multirow{2}{*}{Action Class (\#)} & CTR-GCN & CTR-GCN LSHR (R/I) & Change \\
 & Acc. (\%)&Acc. (\%)&(\%)\\
\hline\hline
Make ok sign (71)  & 41.0 & 50.6 & +9.6 \\
Put something into bag (89)  &  76.0 & 83.48 & +7.5 \\
Cutting paper (scissors) (76)&  61.8 & 69.1 & +7.3 \\
Make victory sign (71) &  35.7 & 40.9 & +5.2 \\
Make a phone call (28) & 84.7 & 89.1 & +4.4\\
Typing on a  keyboard (30) & 64.7& 69.1& +4.4 \\
Sniff/Smell (79) & 77.9 & 82.3 & +4.3 \\
Counting money (74) & 55.6 & 59.6 & +4.0 \\
\hline
%Fold paper (82) & 71.1 & 67.7 & -3.5 \\
%Wield knife towards other person (107) & 73.4 & 69.6 & -3.8 \\
Play magic cube (84) & 71.9 & 67.8 & -4.0 \\
Open bottle (78) & 75.6& 70.7 & -4.9 \\
Blow nose (105) & 64.0 & 67.3 & -5.4 \\
Yawn (103) & 72.7 & 67.3 & -5.4 \\
Reading (11) & 67.0 & 61.2 & -5.9 \\
Staple book (73) & 38.0 & 29.8 & -8.2 \\
\hline
\end{tabular}
\end{center}
\caption{Class accuracy comparison between our method and the CTR-GCN model on NTU120 Cross-Subject using Joint Location using the real and imaginary part.}
\label{table:ctrgcn_ri_quali}
\vspace{-2em}
\end{table}

\begin{table}[h!]
\vspace{2em}
\begin{center}
\begin{tabular}{|l|c|c|c|}
\hline
\multirow{2}{*}{Action Class (\#)} & CTR-GCN & CTR-GCN LSHR (M) & Change \\
 & Acc. (\%)&Acc. (\%)&(\%)\\
\hline\hline
Make victory sign (71) &  35.7 & 46.6 & +11.0 \\
Sneeze/ Cough (41) & 74.3 & 81,2 & +6.9 \\
Typing on a  keyboard (30) & 64.7& 71.3& +6.5 \\
Drop (5) & 82.9 & 89.1 & +6.2 \\
Writing (12) & 54.4 & 60.7 & +6.2 \\
Take off a shoe (17) & 78.1 & 83.2 & +5.1 \\
Make ok sign (71)  & 41.0 & 45.9 & +4.9 \\
Sniff/Smell (79) & 77.9 & 82.4 & +4.5 \\
Point at something with finger (31) & 72.8 & 76.8 & +4.0 \\
\hline
Wipe face \/ feeling warm (37) & 86.6 & 82.6 & -4.0 \\
Use a fan (49) & 95.3 & 90.9 & -4.4 \\
Play magic cube (84) & 71.9 & 67.3 & -4.5 \\
Tennis bat swing (65) & 88.9 & 83.4 & -5.4 \\
Reading (11) & 67.0 & 61.5 & -5.5 \\
\hline
\end{tabular}
\end{center}
\caption{Class accuracy comparison between our method and the CTR-GCN model on NTU120 Cross-Subject using Joint Location using the rotation invariant magnitude.}
\label{table:ctrgcn_mag_quali}
\end{table}

\newpage
%------------------------------------------------------------------------
\section{Classification of action classes}
\label{seca:handclass}

When hand action accuracy is reported in the paper, the mean accuracy for the hand action classes is reported. In preparation, all action classes have been examined as to the extent to which hands are a decisive factor in action detection. The resulting classification of all 120 action classes is made transparent in \autoref{table:action_categorization}.
\begin{table}[h!]
\vspace{-1em}
\begin{center}
\begin{tabular}{|l|l|}
 \hline
Hand action & Body action \\ 
\hline\hline
\multirow{3}{0.36\linewidth}{(A1) drink water, (A2) eat meal/snack,
(A3) brushing teeth, (A4) brushing hair, (A10) clapping, (A11) reading,
(A12) writing, (A13) tear up paper, (A28) make a phone call/answer phone, (A29) playing with phone/tablet, (A30) typing on a keyboard, (A31) pointing to something with finger, (A32) taking a selfie, (A33) check time (from watch), (A34) rub two hands together, (A37) wipe face, (A39) put the palms together, (A49) use a fan (with hand or paper)/feeling warm, (A54) point finger at the other person, (A58) handshaking, (A67) hush (quite), (A68) flick hair, (A69) thumb up, (A70) thumb down, (A71) make ok sign, (A72) make victory sign, (A73) staple book, (A74) counting money, (A75) cutting nails, (A76) cutting paper (using scissors), (A77) snapping fingers, (A78) open bottle, (A81) toss a coin, (A82) fold paper, (A83) ball up paper, (A84) play magic cube, (A85) apply cream on face, (A86) apply cream on hand back, (A89) put something into a bag, (A90) take something out of a bag, (A91) open a box, (A93) shake fist, (A105) blow nose, (A110) shoot at other person with a gun, (A113) cheers and drink, (A115) take a photo of other person, (A112) high-five, (A120) finger-guessing game (playing rock-paper-scissors)} &
\multirow{5}{0.54\linewidth}{(A5) drop, (A6) pickup, (A7) throw, (A8) sitting down,
(A9) standing up (from sitting position), (A14) wear jacket, (A15) take off jacket, (A16) wear a shoe, (A17) take off a shoe, (A18) wear on glasses, (A19) take off glasses, (A20) put on a hat/cap, (A21) take off a hat/cap, (A22) cheer up,(A23) hand waving, (A24) kicking something, (A25) reach into pocket, (A26) hopping (one foot jumping), (A27) jump up, (A35) nod head/bow, (A36) shake head, (A38) salute, (A40) cross hands in front (say stop),(A41) sneeze/cough, (A42) staggering, (A43) falling, (A44) touch head (headache), (A45) touch chest (stomachache/heart pain), (A46) touch back (backache), (A47) touch neck (neckache), (A48) nausea or vomiting condition, (A50) punching/slapping other person, (A51) kicking other person, (A52) pushing other person, (A53) pat on back of other person, (A55) hugging other person, (A56) giving something to other person, (A57) touch other person's pocket, (A59) walking towards each other, (A60) walking apart from each other, (A61) put on headphone,  (A62) take off headphone, (A63) shoot at the basket, (A64) bounce ball, (A65) tennis bat swing, (A66) juggling table tennis balls, (A79) sniff (smell), (A80) squat down, (A87) put on bag, (A88) take off bag,  (A92) move heavy objects, (A94) throw up cap/hat, (A95) hands up (both hands), (A96) cross arms, (A97) arm circles, (A98) arm swings, (A99) running on the spot, (A100) butt kicks (kick backward), (A101) cross toe touch, (A102) side kick, (A103) yawn, (A104) stretch oneself, (A106) hit other person with something, (A107) wield knife towards other person, (A108) knock over other person (hit with body), (A109) grab other person’s stuff, (A111) step on foot, (A114) carry something with other person, (A116) follow other person, (A117) whisper in other person’s ear, (A118) exchange things with other person, (A119) support somebody with hand} \\
& \\
& \\ 
& \\
& \\
& \\
& \\
& \\
& \\
& \\
& \\
 & \\
 & \\
 & \\
 & \\
 & \\
 & \\
 & \\
 & \\
 & \\
 & \\
 & \\
 & \\
 & \\
 & \\
 & \\
 & \\
& \\
& \\
& \\
& \\
& \\
& \\
& \\
& \\
& \\
& \\
& \\
& \\
& \\
& \\
& \\
& \\
& \\
& \\
\hline
\end{tabular}
\end{center}
\caption{Categorization of Action Classes into Hand vs. Non-Hand Actions. Distinction based on action detection focus.} 
\label{table:action_categorization}
\vspace{-2em}
\end{table}
\\
\\

\section{Implementation Details}
\label{seca:impli}
While the main paper describes the characterizing features of the GCN BL and the CTR-GCN model, further implementation details are included in this paragraph.
Both models, i.e.~GCN-BL and CTR-GCN are 10 layers deep and batch normalize the input before the first layer. 
The network output is aggregated over time, joints, and persons, prior to feeding it through first a drop-out layer and then a fully connected layer for action classification using softmax. 
Each graph convolution layer consists of first a spatial GCN module and then a temporal TCN module, which have channel dimensions 64-64-64-64-128-128-128-256-256-256 respectively. 
Both the GCN and TCN modules can be skipped by a residual connection and their output is ReLU activated.
In the GCN-BL model, each spatial GCN unit contains three learnable adjacency matrices $\mathbf{A}$ which are each multiplied with the input  $\mathbf{X}$ and then embedded using a ($1 \times 1$) kernel $\mathbf{W}$. Their embeddings are summed up and batch normalized, i.e. $\mathbf{Z} = \sum_{|A_{dim}|} \mathbf{W} (\mathbf{X} \mathbf{A})$. This model uses the learned adjacency matrices $\mathbf{A}$ to compute a static shared topology. A residual connection allows skipping these steps. Finally, a ReLU activation is applied. The temporal relations between the joints are assessed in the TCN unit by using a ($5 \times 1$) convolution on each individual joint coordinate which spans five subsequent time steps. Further, batch normalization is applied. The CTR-GCN's GCN module contains three CTR-GC units instead of adjacency matrix multiplications. 
They create an embedding $\mathbf{Z} = \mathcal{A}(\mathbf{X}\mathbf{T}, \mathcal{R} (\mathcal{M}, \mathbf{X}), \mathbf{A}))$ based on the input data and a learned graph topology.
Each CTR-GC contains two branches, both starting with a ($1 \times 1$) convolution to reduce channel dimensionality. 
Further, temporal pooling is applied prior to a pair-wise subtraction of the output of the two branches and an activation, defined as $\mathbf{T}\in  \mathbb{R}^{ N \times C \times C'}$.
The graph adjacency matrix $\mathbf{A} \in  \mathbb{R}^{ N \times V \times V}$ is used to refine the learned channel-wise topologies $\mathcal{M} \in  \mathbb{R}^{ N \times V \times V} $. 
Lastly, channel-wise aggregation is computed using function $\mathcal{A}$. The CTR-GCs' output is summed, batch normalized, and ReLU activated; the entire GCN unit can be skipped using a residual connection. 
The TCN module contains four branches of convolutions, each with different temporal kernels and dilations, to account for various action lengths, whose output is concatenated.
\newpage
\newpage
\newpage

\end{document}